# Early Discovery of Emerging Entities in Persian Twitter with Semantic Similarity

Shahin Yousefi
Dept. of Computer Science and Information Technology,
Institute for Advanced Studies in Basic Sciences (IASBS),
Zanjan, Iran
sh.yousefi@iasbs.ac.ir

Mohsen Hooshmand
Dept. of Computer Science and Information Technology,
Institute for Advanced Studies in Basic Sciences (IASBS),
Zanjan, Iran
Mohsen.hooshmand@iasbs.ac.ir

Mohsen Afsharchi
Dept. of Electrical & Computer Engineering,
University of Zanjan,
Zanjan, Iran
afsharchi@znu.ac.ir

*Abstract*—Discovering *emerging entities* (EEs) is the problem of finding entities before their establishment. These entities can be critical for individuals, companies, and governments. Many of these entities can be discovered on social media platforms, e.g. Twitter. These identities have been the spot of research in academy and industry in recent years. Similar to any machine learning problem, data availability is one of the major challenges in this problem. This paper proposes EEPT. That is an online clustering method able to discover EEs without any need for training on a dataset. Additionally, due to the lack of a proper evaluation metric, this paper uses a new metric to evaluate the results. The results show that EEPT is promising and finds significant entities before their establishment.

*Keywords—Emerging Entity; Semantic Similarity; Online Clustering; Text Clustering; Early Discovery*

## I. INTRODUCTION

Nowadays, having an information advantage is critical for governments, organizations, and even individuals. The institutes use the information for better analysis, decision-making, and action to have higher throughput or profit. One aspect of this advantage is being informed of an entity before its establishment. An entity refers to a person, event, object, or date and generally, it is a proper noun [7, 13]. An emerging entity, or EE, is a concept or entity that has been increasingly referred to or discussed either in the news or on social media and is now often referenced in a single article or source [10, 11]. The tracking of emerging entities such as new products, jobs, and individuals is a beneficial task for numerous applications, including social-trend analysis and marketing research. To this end, real-time monitoring of emerging entities is a must for the successful implementation of such applications. A knowledge Base, or KB, is a structured repository for storing complex data [8]. KBs such as Wikipedia, are a source of information, though they tend to be slow in registering new entities, and only those of particular note are chosen for inclusion [10][12]. On the other hand, social media can provide early identification of emerging entities prior to their formal establishment, allowing a more comprehensive scope of emerging entities than other traditional sources, such as newspapers, television, and news agencies [11]. Therefore, attending to these media, such as Twitter, can be beneficial for identifying these entities.

Previous research has focused on identifying entities that are not present in a given KB, leading to the discovery of a large number of non-emerging entities, as the absence of such entities in a KB does not guarantee their emergence [3, 4, 5, 6]. Additionally, using KB alone will not account for homographic entities, which are words with the same spelling but different meanings. In another research conducted by [10], the use of emerging contexts to extract emerging entities was explored. However, due to the potential lack of clarity in the context of the potential unavailability of the requisite dataset to facilitate model training, the full potential of this approach cannot be realized. The study [2] utilized Twitter trends to identify potential emerging entities. As our results in section IV indicate, Twitter trends do not provide comprehensive coverage of all emerging entities and may include non-emerging entities.

In this paper, we propose an online clustering method that is able to discover emerging entities without any need for training on a dataset. First, the data is collected from Twitter and then clustered using an online clustering algorithm called textClust [9]. This algorithm weights and clusters the tweets according to their importance and frequency, and removes data that become insignificant over time. Then, using a threshold, the words and phrases from the clusters of higher weight (e.g., containing a higher number of tweets) are extracted and checked against the Wikipedia knowledge base. If the extracted words and phrases are not found in the knowledge base, they are introduced as emerging entities. This online clustering is important for finding similar entities together. In addition to the weight calculation, a pre-trained fastText model is used to calculate the semantic similarity between the data. The proposed method is also able to utilize time intervals to discover emerging entities as quickly as possible. To evaluate the efficacy of the proposed method, the data was collected and evaluated over the course of two months. The results demonstrate that the proposed method is more effective and efficient in discovering emerging entities than the Twitter trend.

Our contributions are as follows:

- Online Clustering: developing a novel algorithm to solve the problem of discovering emerging entities in an online setting.
- Semantic Similarity: proposing a method to identify similar entities through semantic similarity.
- Early Discovery: introducing a technique to enable the early detection of entities utilizing time intervals.
- No Training Data: the model was designed to produce superior results without the need for training data while allowing for testing data to remain unlabeled.

This paper presents a method, EEPT, for addressing the emerging entity problem. In Section II, reviews the existing methods used to address the emerging entity problem. Section III proposes EEPT as our proposed method, along with a baseline. The data and implementation details of EEPT are reported in Section IV, along with a specific performance metric for the unlabeled data. The results and performance of EEPT are also discussed in this section. Finally, Section V provides a conclusion.

## II. RELATED WORK

There are some studies are available on emerging entities. Previous studies have examined the identification of entities that are not accounted for in a given knowledge base (KB), resulting in the identification of numerous non-emerging entities as the lack of presence of such entities in a KB does not guarantee their emergence. Reference [6] proposed a method of utilizing NER to extract Named Entities (NEs), classifying all non-registered NEs as out-of-KB. However, this technique is not suitable for homographs, as it disregards the contexts in which these NEs are presented. In [3, 4, 5] proposed approaches for determining whether a given NE in a news article is out-of-KB. However, their approaches are not suitable for the problem of dealing with the ever-increasing number of newly emerging entities.

In [10], the application of emergent contexts for the identification of emerging entities was investigated. However, the effectiveness of this approach is limited by the potential for inadequate context clarity, as well as the possible unavailability of the necessary dataset for effective model training. To leverage the full potential of such an approach, further study is needed to evaluate the impact of data availability, context specificity, and other factors on the capability of the model to effectively detect emergent entities.

The research conducted by [2] incorporated the use of Twitter trends to detect potential burgeoning entities. However, it should be noted that Twitter trends are not a reliable source for identifying all possible rising entities, as they may also contain non-emerging entities.

Generally, these studies suggest that relying solely on a knowledge base is insufficient for detecting emerging entities. Furthermore, utilizing Twitter trends to identify emerging entities is not a viable option. Additionally, due to the particular challenges associated with Persian Twitter (such as ironic, metaphorical, too short sentences, etc.), using emerging content is not a feasible solution.

## III. PROPOSED METHOD

In this section, we describe proposed method to discover emerging entities in Persian Twitter called EEPT. We use Twitter since [11] indicated that a broad selection of emerging entities arises earlier on social media. EEPT performs an analysis of tweets to identify potential emerging entities. As we show in section IV, the proposed method outperforms the Twitter trends.

EEPT uses online clustering textClust [9] as a baseline. The idea of online clustering is important for finding similar entities together. Our method does not need any data for training and the test data also are unlabeled.

First, we elucidate the approach of the textClust algorithm. The textClust is an algorithm that uses cosine similarity to identify similarities between documents and cluster them accordingly. It can be used for tasks such as retrieval, summarization, and classification. The algorithm tokenizes the text, then calculates its n-gram, and assigns a temporary micro-cluster to it. Subsequently, the cosine similarity between the TF-IDF vector of the temporary micro-cluster and all other micro-clusters is measured in order to identify the closest micro-cluster to it. If the nearest micro-cluster is within the predetermined threshold, it is merged with it; otherwise, the temporary micro-cluster is added to the model as a new micro-cluster. The fading function is utilized to erase obsolete clusters, which helps in identifying emerging entities over a certain period and filtering out noisy items. This is a great advantage of the algorithm as it allows for the efficient detection of emerging entities. Finally, micro-clusters are clustered to form macro-clusters or final clusters.

Figure 1 illustrates the overall structure of the Enhanced EEPT. First, the data taken from Twitter was segmented. Then, the tweets were processed by the textClust algorithm which included tokenizing the tweet and creating related n-grams and a temporary microcluster. Subsequently, the semantic similarity between tweets was calculated using the pre-trained fastText [1] model which is embedded for the Persian language. This technique enabled the detection of similar emerging entities together. After that, the presence or absence of the extracted words in Wikipedia was checked as the knowledge base and the phrases that are not registered in Wikipedia were identified as emerging entities in the output.

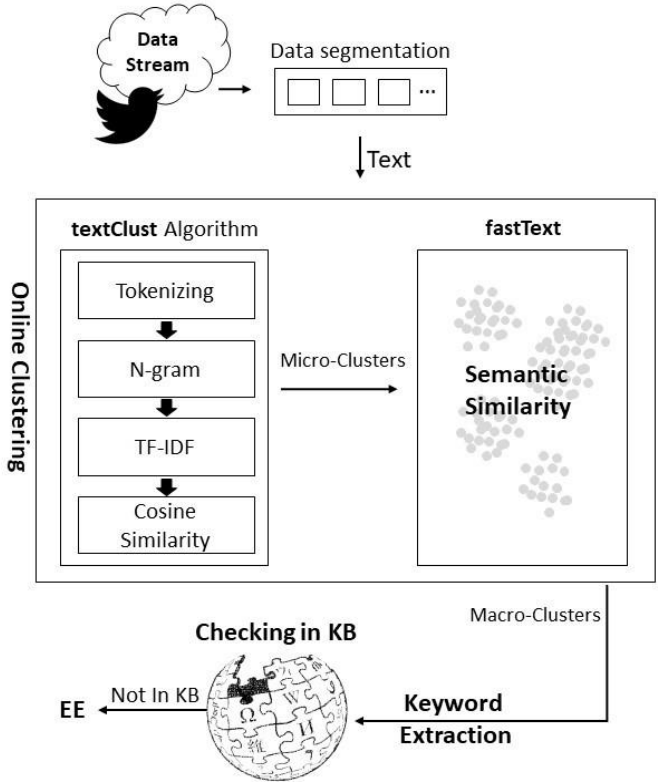

Fig. 1. The general framework of the presented method.

The concept of data segmentation aligns with the early discovery of appearing entities. By doing this, we can execute the entire algorithm for every time interval, and if entities are manifesting in that time interval, we can extract them quickly.

TABLE I. RESULTS OF APPLYING EEPT ON COLLECTED DATA.

| ID | Date-Intervals | No. of EE found by Twitter | No. of EE found by EEPT | Acc$^{nl}$ |
|---|---|---|---|---|
| 1 | Oct. 22-Oct. 25 | 2 | 1 | 0.5 |
| 2* | Oct. 26-Oct. 29 | 2 | 2 | 0.33 |
| 3* | Oct. 30-Nov. 02 | 2 | 1 | 0 |
| 4* | Nov. 03-Nov. 06 | 1 | 2 | 0.5 |
| 5 | Nov. 07-Nov. 10 | 1 | 1 | 1 |
| 6 | Nov. 11-Nov. 14 | 1 | 1 | 1 |
| 7 | Nov. 21-Nov. 24 | 2 | 1 | 0.5 |
| 8 | Nov. 25-Nov. 28 | 4 | 4 | 1 |
| 9 | Nov. 29-Dec. 02 | 1 | 1 | 1 |
| 10* | Dec. 03-Dec. 06 | 1 | 1 | 0 |
| 11* | Dec. 07-Dec. 10 | 3 | 2 | 0.25 |
| 12 | Dec. 11-Dec. 14 | 2 | 2 | 1 |

## IV. EVALUATION

In this section, we present the results of applying our algorithm to Twitter data to identify emerging entities. Early discovery of emerging entities requires an accessible, up-to-date corpus of media. As [11] noted, a greater number and wider range of emerging entities can be found on social media and micro-blogs before being reported by news agencies. Such platforms can provide the fastest means of identifying emerging entities. Twitter is one of the most popular social media platforms in the world with this capability. In this paper, we utilize Wikipedia as our knowledge base (KB). As mentioned in [10], registration of entities in KBs such as Wikipedia is often delayed, making them suitable candidates for emerging entities. Moreover, Wikipedia is the most comprehensive online KB in many languages, including Persian [11].

### A. Data collection

Using API, we collected Persian Twitter data from October 22$^{nd}$ to December 14$^{th}$ 2021. Each data-point is assumed as a triple (*id*; *t*; *ts*), such that *id* is a unique value as the identification number; *t* is the text of the tweet; *ts* is the timestamp of the tweet including its arrival date and time. Parameter *ts* is crucial, as it allows for the gradual fading of outdated data over time. The collected data contains 453k tweets. Due to incomplete data collection on dates between November 15$^{th}$ and November 20$^{th}$, we removed their corresponding data from the collected data. Then, we segmented data into four-day intervals.

### B. Pre-processing

In the pre-processing phase of the collected data, punctuation marks, spaces, URLs, and emojis were removed. Stemming was not necessary to ensure the retrieval of exact phrases. Numbers were not removed as some of the emerging entities include numbers, e.g. COVID-19.

### C. Implementation

We divided data into four-day intervals for implementation purposes. We set the n-grams to a minimum and maximum of bigram and trigram, respectively, due to most of the unigrams being present in the knowledge base (KB). We identified macro-clusters with more than 10k tweets and extracted the five top-ranked phrases by keyword extraction from each chosen cluster. The EEPT then checked these phrases against Wikipedia; if not available, it labeled them as emerging entities.

### D. Metric for unlabeled dataset

In this work, we present a new evaluation criterion for assessing the test data with no labeling, given that our model does not require training data, and its implementation and testing are not similar to previous works. This evaluation criterion takes into account True Positive (TP), False Positive (FP), and False Negative (FN) values and calculates as follows:

$$Acc^{nl} = \frac{TP}{TP + FP + FN} \quad (1)$$

We use $Acc^{nl}$ to evaluate the EEPT for finding emerging entities.

*E. Experimental Results*

In this section, we report the simulation of the EEPT. We run the algorithm on the dataset segmented into four-day intervals.

In Section III, the proposed method for processing tweets was explained in detail. Figure 2 illustrates the output of the proposed method for the time interval of October 22[nd] to October 25[th], which can be divided into two parts: the left side displays the output of the online clustering procedure, and the right side displays the keywords extracted from the clusters that met the desired threshold.

Figure 2 demonstrates the clustering of tweets according to their content. The x-axis represents the number of tweets, while the y-axis displays the number of clusters. Clusters that are lighter in color and larger in size indicate a higher number of tweets with similar content. We set a threshold of 10k tweets to identify cluster words. The right side of Figure 2 shows the output of the yellow cluster words, as a Word Cloud, with the size of each word reflecting its importance (We omit the exact expressions from the figure and replace them with "phrase + number" to increase readability. The exact Persian phrases are provided in Figure 3).

Table I presents the simulation results. Each row contains four parameters, including an ID of the interval, the corresponding date interval, the number of emerging entities found by the Twitter trend, the number of emerging entities found by the EEPT, and the accuracy of the method regarding the emerging entities ($Acc^{nl}$). The accuracy of the data, as shown in the last column of the table, is between 0 and 1. The results obtained demonstrate that the desired accuracy was achieved. Some of the IDs were marked with a "*", which showed interesting behavior. Upon further analysis, it was revealed that EEPT was able to detect entities that the Twitter trend was unable to report not even after registering them. However, it is possible that the accuracy of EEPT is lower in these intervals. This demonstrates the power of EEPT.

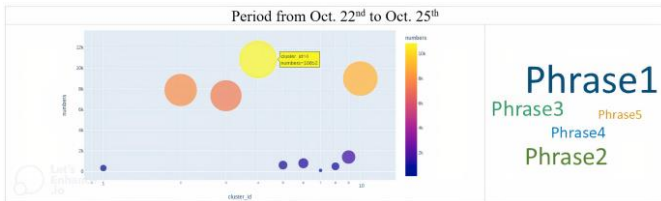

Fig. 2. The results of the proposed method for the period between October 22[nd] and October 25[th].

## V. Conclusion

This paper proposed a method for the early discovery of emerging entities in Persian Twitter, named EEPT. To semantically cluster the unlabeled data, we employed the textClust algorithm and fastText. Additionally, time intervals were utilized to discover emerging entities in advance. Moreover, the semantic similarity was used to extract similar entities together. We also proposed a novel metric for evaluating our method, which was used to compare our model with the Twitter trend. As a result, our model outperformed the Twitter trend. However, due to the unavailability of a Persian Emerging Entity dataset, we faced a challenge in evaluating our model. In the future, it is necessary to produce a labeled dataset to enable better methods and analysis.

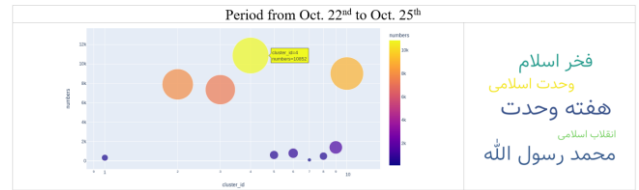

Fig. 3. The exact phrases in Persian found for October 22[nd] and October 25[th]